\documentclass[sigconf]{acmart}

\AtBeginDocument{%
  \providecommand\BibTeX{{%
    \normalfont B\kern-0.5em{\scshape i\kern-0.25em b}\kern-0.8em\TeX}}}

\usepackage[normalem]{ulem}
\usepackage{booktabs} 
\usepackage{subcaption}

\usepackage[english]{babel}
\usepackage[utf8x]{inputenc}
\usepackage[T1]{fontenc}
\usepackage{comment} 

\usepackage{amsmath}
\usepackage{graphicx}
\usepackage{xcolor}
\usepackage[colorinlistoftodos]{todonotes}
\usepackage{multirow}

\copyrightyear{2021}
\acmYear{2021}
\setcopyright{acmcopyright}\acmConference[GECCO '21]{2021 Genetic and Evolutionary Computation Conference}{July 10--14, 2021}{Lille, France}
\acmBooktitle{2021 Genetic and Evolutionary Computation Conference (GECCO '21), July 10--14, 2021, Lille, France}
\acmPrice{15.00}
\acmDOI{10.1145/3449639.3459323}
\acmISBN{978-1-4503-8350-9/21/07}

\begin{document}

\title[Using Multiple GANs to Build Better-Connected Levels for Mega Man]{Using Multiple Generative Adversarial Networks to Build\\ Better-Connected Levels for Mega Man}


\author{Benjamin Capps}
\affiliation{%
  \institution{Southwestern University}
  \streetaddress{1001 E. University Ave}
  \city{Georgetown} 
  \state{Texas}
  \country{USA}
  \postcode{78626}
}
\email{cappsb@southwestern.edu}

\author{Jacob Schrum}
\orcid{0000-0002-7315-0515}
\affiliation{%
  \institution{Southwestern University}
  \streetaddress{1001 E. University Ave}
  \city{Georgetown} 
  \state{Texas}
  \country{USA} 
  \postcode{78626}
}
\email{schrum2@southwestern.edu}

\begin{abstract}
Generative Adversarial Networks (GANs) can generate levels for a variety of games. This paper focuses on 
combining GAN-generated segments in a snaking pattern to create levels for Mega Man. Adjacent segments in such levels can be orthogonally adjacent in any direction, meaning that an otherwise fine segment might impose a barrier between its neighbor depending on what sorts of segments in the training set are being most closely emulated: horizontal, vertical, or corner segments. To pick appropriate segments, multiple GANs were trained on different types of segments to ensure better flow between segments. Flow was further improved by evolving the latent vectors for the segments being joined in the level to maximize the length of the level's solution path. Using multiple GANs to represent different types of segments results in significantly longer solution paths than using one GAN for all segment types, and a human subject study 
verifies that these levels are more fun and have more human-like design than levels produced by one GAN.
\end{abstract}


%
%



\maketitle

\section{Introduction}
Generative Adversarial Networks (GANs \cite{goodfellow2014generative}) are capable of reproducing certain aspects of a given training set. GANs are artificial neural networks that can be trained to generate fake samples based on real examples. Past successes include the generation of fake celebrity faces~\cite{karras:iclr2018} and fingerprints~\cite{bontrager2017deepmasterprint}. In the domain of games, GANs have generated levels for Mario and others~\cite{volz:gecco2018,gutierrez2020zeldagan,giacomello:cog19,torrado:cog20}. 

In this paper, GANs are used to generate Mega Man levels based on levels in the original game. Data from the Video Game Level Corpus (VGLC \cite{summerville:vglc2016}) is used to train GANs. Volz et al.~\cite{volz:gecco2018} generated Mario levels by placing individual segments left-to-right. In contrast to Mario, Mega Man levels have a snaking pattern of horizontal, vertical, and corner segments. 
Therefore, different GANs are used for different segment types, resulting in levels with better flow and organization, and a more human-like design. 

Levels were optimized using latent variable evolution (LVE~\cite{bontrager2018deep}): one real-valued vector consisting of the concatenation of multiple latent vectors was used to generate several segments. The vector also contained information on the relative placement of each segment. Levels were evolved using multiple objectives by Non-Dominated Sorting Genetic Algorithm II (NSGA-II~\cite{deb:tec02}), the most relevant being
solution path length as determined by A* Search. 


Our new approach, MultiGAN, trains distinct GANs on different portions of the training data, and queries the appropriate GAN for each segment type as needed.
MultiGAN is compared with the standard approach of training one GAN on all data: OneGAN.
Although segments produced by GANs make no assumptions regarding neighboring segments,
LVE encourages sensible transitions between segments. However, MultiGAN more easily selects appropriate segments, resulting in significantly longer solution paths, and levels that flow in a more human-like fashion. In contrast, OneGAN levels are chaotic and have shorter solutions, since irregular boundaries and undesirable shortcuts emerge despite evolution.

A human subject study was also conducted
that indicates MultiGAN levels are significantly more fun and human-like in their design than OneGAN levels, confirming our analysis of the levels, and indicating that MultiGAN is a promising approach for generating better levels for classic platforming games.


\section{Related Work}
\label{sec:related}

Procedural Content Generation (PCG~\cite{togelius2011procedural}) is an automated way of creating content for video games. PCG via Machine Learning (PCGML~\cite{Liu_2020}) is a way of combining machine learning with PCG.

Generative Adversarial Networks are an increasingly popular PCGML method for video game level generation. After the GAN is properly trained, randomly sampling vectors from the induced latent space generally produces outputs that could pass as real segments of levels. However, some segments are more convincing than others, or have other desirable qualities (e.g.\ enemy count, solution length, tile distribution), so it makes sense to search the latent space for these desirable segments via methods such as evolution.


The first latent variable evolution (LVE) approach was introduced by Bontrager et al.~\cite{bontrager2017deepmasterprint}. In their work a GAN was trained on a set of real fingerprint images and then evolutionary search was used to find latent vectors that matched subjects in the data set. 
Because the GAN is trained on a specific target domain, most generated phenotypes resemble valid domain artifacts.
The first LVE approach to generating video game levels was applied to Mario~\cite{volz:gecco2018}. Work quickly followed in other domains.


Giacomello et al.~\cite{giacomello:cog19} used a GAN to generate 3D levels for the First-Person Shooter Doom. 
Gutierrez and Schrum \cite{gutierrez2020zeldagan} used a Graph Grammar in tandem with a GAN to generate dungeons for The Legend of Zelda. Work in the GVG-AI \cite{GVGAI:TCIAIG2016} variant of Zelda was also done by Torrado et al.~\cite{torrado:cog20} using conditional GANs. This work used bootstrapping to help in the generation of more solvable levels. A similar bootstrapping approach was also used by Park et al.~\cite{park:cog19} in an educational puzzle-solving game.
Additional work has also been done in a broader collection of GVG-AI games by Kumaran et al.~\cite{kumaran:aiide2020}, who used one branched generator to create levels for multiple games derived from a shared latent space.




Additional work has also been done in the original Mario domain.
In particular, Fontaine et al.~\cite{fontaine2020illuminating} and Schrum et al.~\cite{schrum:gecco2020cppn2gan} have both used the quality diversity algorithm MAP-Elites \cite{mouret:arxiv15} to find a diversity of interesting Mario levels. The approach by Schrum et al.\ specifically used Compositional Pattern Producing Networks \cite{stanley:gpem2007} with GANs to make levels with better cohesion and connectivity. This approach was applied to Zelda as well as Mario. Schrum et al.~\cite{schrum2020interactive} also combined GANs with interactive evolution in these domains to search the space of levels according to user preferences.



Previous work with GANs in Mario \cite{volz:gecco2018,fontaine2020illuminating,schrum:gecco2020cppn2gan,schrum2020interactive} all learn to generate one level segment at a time, before placing the adjacent segments left-to-right. Mega Man levels are more complicated in that they have a snaking-pattern that can move right, up, down, and even to the left. The only other paper that has addressed this challenge is recent work by Sarkar and Cooper \cite{sarkar:fdg2020} that uses Variational Autoencoders (VAEs). This method was applied to horizontal Mario levels, vertical Kid Icarus levels, and snaking Mega Man levels. Although their results seem promising, they often still have problems with barriers between segments. In their work, A* paths through the training levels were part of training data, and proposed paths are actually part of the VAE output. However, these paths are not always valid, and example levels shown in the paper are not always traversable. Therefore, in our work, levels are specifically optimized to maximize the resulting A* path length, to assure that the Mega Man levels are actually traversable.


\begin{table}[t!]
\caption{\label{tab:tiles}Tile Types Used in Mega Man.}
{\small Tile types come from VGLC, though additional types were added based on observations of the actual game. The original VGLC did not include enemies, level orbs, or water. The ``In VGLC'' column indicates whether the tile was originally represented in VGLC. ``Training'' indicates whether the tile was used in GAN training sets. ``Char'' is the original character code representation in VGLC (or a made up code for tiles not in VGLC), and ``Int'' is a numeric code used in JSON representations of the training data. In VGLC, one tile was associated with a Cannon obstacle. That tile maps to Int 6, but was deemed unnecessary and becomes a solid block in generated levels. Additionally, although no specific enemy was used for training, a single general enemy type with code 11 was used as a placeholder for an enemy, and an algorithm later specified the type based on location.}

\centering
\begin{tabular}{|c|c|c|c|c|c|}
\hline
Tile type & In VGLC & Training & Char& Int & Game \\ 
\hline 
Air & Yes & Yes & \texttt - & 0 & \raisebox{-.275\height}{\includegraphics[width=.05\columnwidth]{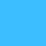}} \\
\hline
Solid & Yes & Yes & \texttt \# & 1, 6 & \raisebox{-.275\height}{\includegraphics[width=.05\columnwidth]{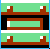}} \\
\hline
Ladder & Yes & Yes &\texttt | & 2 & \raisebox{-.275\height}{\includegraphics[width=.05\columnwidth]{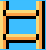}} \\
\hline
Hazard & Yes & Yes &\texttt H & 3 & \raisebox{-.275\height}{\includegraphics[width=.05\columnwidth]{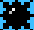}} \\
\hline
Breakable & Yes & Yes  &\texttt B & 4 & \raisebox{-.275\height}{\includegraphics[width=.05\columnwidth]{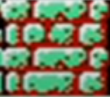}} \\
\hline
Moving  Platform & Yes & Yes &\texttt M & 5 & \raisebox{-.275\height}{\includegraphics[width=.05\columnwidth]{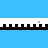}} \\
\hline
Orb & No & No &\texttt Z & 7 & \raisebox{-.275\height}{\includegraphics[width=.05\columnwidth]{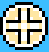}} \\
\hline
Player & Yes & No &\texttt P & 8 & \raisebox{-.275\height}{\includegraphics[width=.05\columnwidth]{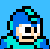}} \\
\hline
Null & Yes & Yes &\texttt @ & 9 & \raisebox{-.275\height}{\includegraphics[width=.05\columnwidth]{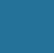}} \\
\hline
Water & No & Yes &\texttt W & 10 & \raisebox{-.275\height}{\includegraphics[width=.05\columnwidth]{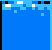}} \\
\hline
Generic Enemy & No & Yes &\texttt Varies & 11 & N/A \\
\hline
Ground Enemy & No & No &\texttt G & 11 & \raisebox{-.275\height}{\includegraphics[width=.05\columnwidth]{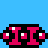}} \\
\hline
Wall Enemy & No & No &\texttt W & 12 & \raisebox{-.275\height}{\includegraphics[width=.05\columnwidth]{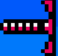}} \\
\hline
Flying Enemy & No & No &\texttt F & 13 & \raisebox{-.275\height}{\includegraphics[width=.05\columnwidth]{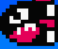}} \\
\hline

\end{tabular}
\end{table}

\section{Mega Man}


Mega Man (Rockman in Japan), was released in 1987 on the Nintendo Entertainment System (NES). Gameplay involves jumping puzzles, killing enemies, and a boss at the end of each level. Mega Man's success led to numerous sequels on the NES and other systems, most with the same graphics aesthetic and game mechanics.


The Video Game Level Corpus (VGLC \cite{summerville:vglc2016}) contains data from numerous games, including Mega Man. VGLC is the source of training data for 
much of the research in Section \ref{sec:related} \cite{volz:gecco2018,gutierrez2020zeldagan,giacomello:cog19,sarkar:fdg2020,schrum:gecco2020cppn2gan,schrum2020interactive}. The levels are represented as text files, where each character represents a different tile type from the level, as seen in 
Table~\ref{tab:tiles}. The original VGLC data is slightly modified and enhanced as described in the table caption. Additionally, a level orb is added to mark the end of each level (bosses not included).
The snaking pattern of some levels presents an interesting challenge for level generation. 


Mega Man Maker\footnote{\url{https://megamanmaker.com/}} is a fan-made game for building and playing user-created levels. The game includes content from each Mega Man game, including a platform-gun that allows the player to traverse otherwise difficult jumping puzzles with more ease. Mega Man Maker is used to visualize and play levels generated by the GANs.


\section{Approach}

First, data was collected from VGLC for training the GANs. OneGAN was trained in a manner similar to previous PCGML work with GANs, but MultiGAN required the collected training data to be separated by type. However, levels were evolved for both approaches using the same genome encoding.


\begin{figure}[t]
\centering
\includegraphics[width=1.0\columnwidth]{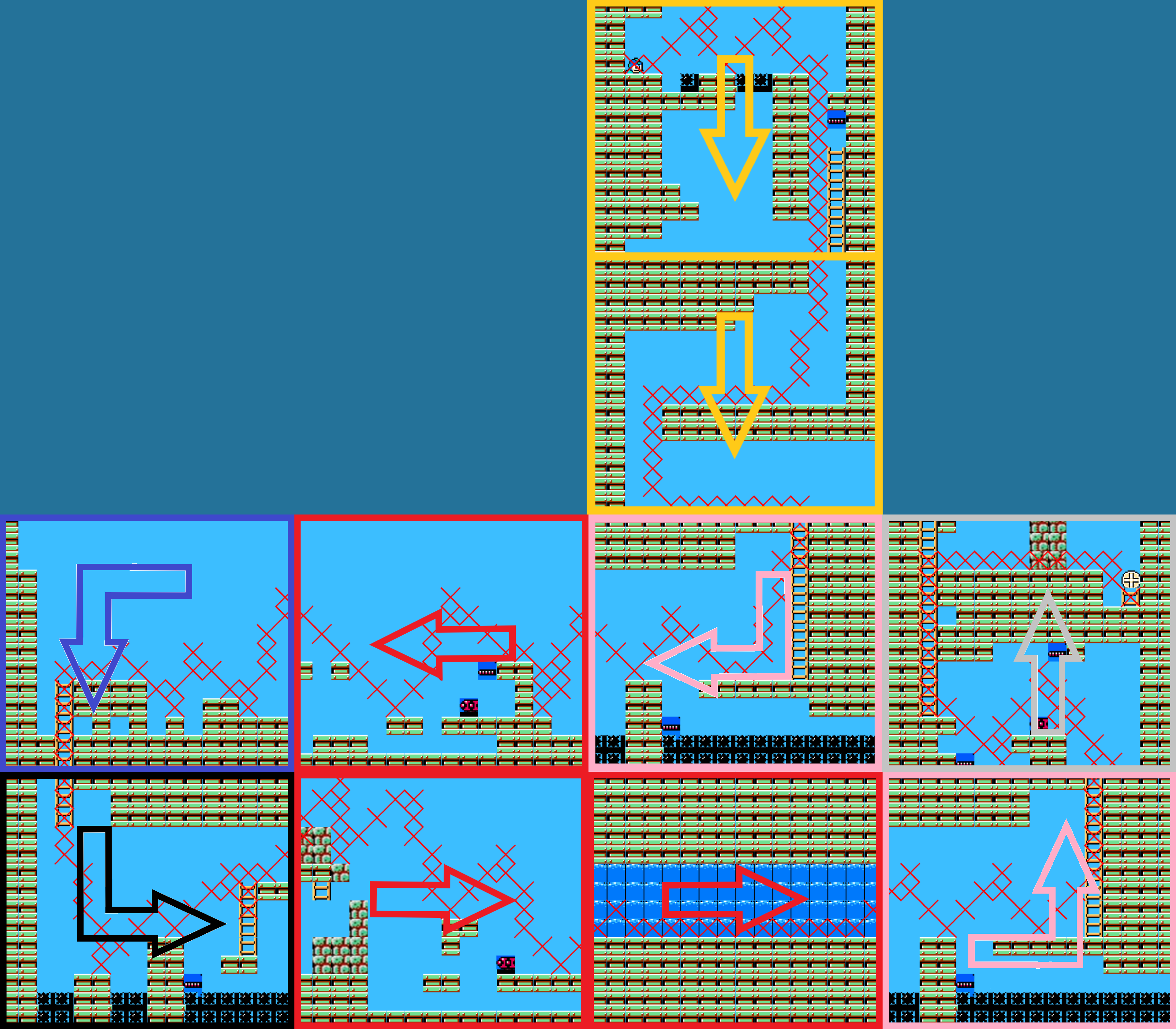}
\caption{Level Generated by MultiGAN. \normalfont Each color represents a segment of a different type: Yellow:Down, Pink:Lower Right, Red:Horizontal, Blue:Upper Left, Black:Lower Left, White:Up. Upper Right segments can also be generated, but none are shown. Each GAN was trained only on the data of that given type.}
\label{fig:MultiGANmodel}
\end{figure}


\subsection{Collecting Data}


MultiGAN requires the training data to be categorized by type:
up, down, horizontal, lower left, lower right, upper left, and upper right segments (Fig.~\ref{fig:MultiGANmodel}). Up and down segments are distinct because down segments often involve falling, thus making it impossible to move upward. Up segments require ladders and/or platforms that enable upward movement. In contrast, there is no distinction between left and right segments, which are both horizontal.


The VGLC data was missing many details from the original game, some of which were added back to the data before training (Table~\ref{tab:tiles}). 
In contrast, player spawn points were replaced with air tiles. Only one is needed per level, which is better placed using simple rules.

The modified level files were scanned with a sliding window to create training data. The window size corresponds to the area visible on screen, which is 14 tiles high by 16 tiles wide. The window slides one tile at a time in the appropriate direction given where the path of the level led next. 
Although adjacent segments can be orthogonally adjacent in any direction, there is no branching, so there is always exactly one direction to head in toward the end of the level. 
Horizontal, up, and down segments are categorized according to the direction the window slides while collecting the data. A segment is identified as a corner segment whenever the direction of sliding has to change. However, each corner segment is also considered 
a horizontal, up, or down segment, depending on which direction the window slides when entering the segment.
The result of this process is a collection of $14 \times 16$ segment training samples. OneGAN is trained with all of the data, whereas MultiGAN had a separate training set for each segment type (Table~\ref{tab:GANs}).



\begin{table}[t]
\caption{\label{tab:GANs}Characterization of VGLC Mega Man Data.}
{\small Number of segments of each type in the training data. OneGAN
 uses all training data, but MultiGAN uses a separate training set for each of seven distinct GANs, each consisting of samples with the same segment type.}

\centering
\begin{tabular}{|c|c|c|}
\hline
GAN & Type & Segment Count \\ 
\hline 
OneGAN & All & 2344 \\
\hline
\multirow{7}{*}{MultiGAN} & Horizontal & 1462 \\
\cline{2-3}
& Up & 518  \\
\cline{2-3}
& Down & 364 \\
\cline{2-3}
& Upper Left Corner & 10   \\
\cline{2-3}
& Lower Right Corner & 9  \\
\cline{2-3}
& Upper Right Corner & 8 \\
\cline{2-3}
& Lower Left Corner & 8 \\
\hline
\end{tabular}
\end{table}

\subsection{Training the GANs}

\begin{figure}[t]
\centering
\includegraphics[width=\columnwidth]{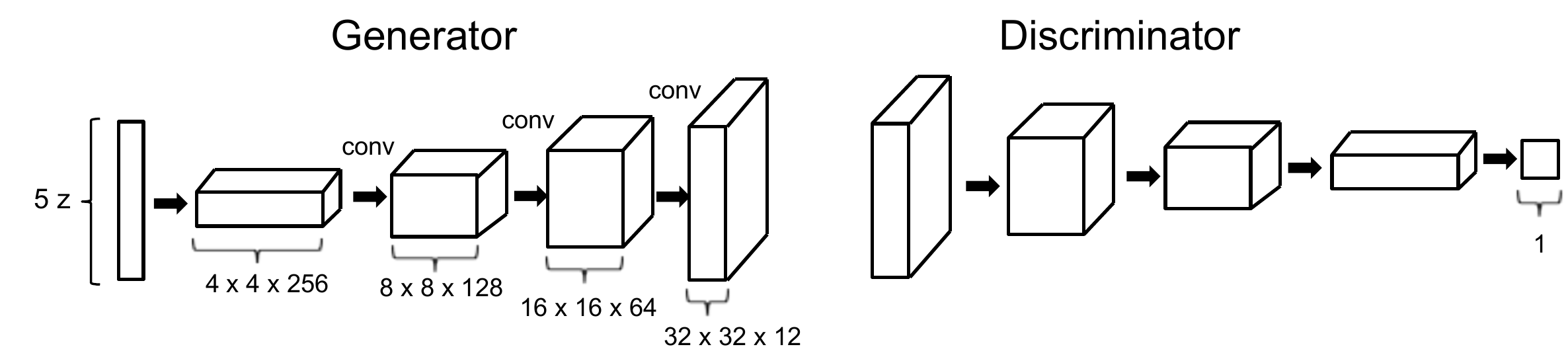}
\caption{GAN architecture.}
\label{fig:architecture}
\end{figure}

The type of GAN used is a Deep Convolutional GAN, specifically a Wasserstein GAN \cite{arjovsky2017wasserstein}, as used in previous studies \cite{volz:gecco2018,gutierrez2020zeldagan,schrum:gecco2020cppn2gan}. 
The specific architecture is shown in Fig.~\ref{fig:architecture}.

There are two key components to training a GAN: a generator and a discriminator. The generator is the GAN itself, and is responsible for generating fake outputs based on an input latent vector. The discriminator is trained to recognize whether a given input is real (from the training set) or fake (from the generator). 

The 2D JSON training segments of size\footnote{The actual size of the generator output and discriminator input is $32 \times 32$ for compatibility with previous research. Inputs are padded with zeros to be the right size.} $14 \times 16$ are expanded into 3D volumes by converting each integer into a one-hot vector across $12$ channels, one per tile type in Mega Man. During training, the discriminator is shown real and fake level segments. 
Fake segments are generated by the GAN by giving it
randomized latent vectors of length $5$.
Discriminator weights are adjusted to be more accurate, and generator weights are adjusted to produce better fakes.
The GAN is trained for 5000 epochs, at which point the discriminator can no longer determine whether an image is real or fake, and is thus discarded.
However, the generator can now produce convincing fake level segments
given arbitrary latent vector inputs.







The OneGAN is trained on all data, which is the norm, but as a result there is no clear way to retrieve a segment of a desired type.
MultiGAN is trained on the same data, but each individual GAN of the MultiGAN was trained on a different category of data from Table~\ref{tab:GANs}. Each of the seven GANs had the same architecture (Fig.~\ref{fig:architecture}) and therefore same latent input size of 5 as OneGAN. Each was also trained for 5000 epochs.


\subsection{Genome Encoding}
\label{sec:genome}



Complete levels are generated from MultiGAN and OneGAN in the same way. 
The genome is a vector of real numbers from $[-1,1]$ divided into sections for each segment. In each section of 9 variables, the first 5 are latent variables, and the last 4 determine the relative placement of the next segment. These 4 values correspond to up, down, left, right placement. Whichever direction has the maximum value is chosen for the next placement, unless that location is occupied, in which case the next highest value is chosen, and so on until an unoccupied neighboring location is found. If there are no unoccupied neighbors, then level generation terminates early.

For each segment generated by OneGAN, the 5 latent variables of the genome section are sent to the GAN to produce the segment. For MultiGAN, the direction of the next placement and the previous placement determine the appropriate type for the current segment, and the specific GAN for the needed type is used to generate the segment. If the new direction is different from the previous one, a corner GAN is used to generate that segment. If a segment is generated to the right, and the new direction will be up, then the lower-right GAN will generate a segment to the right to prepare for the new up segment, and so on.



For simplicity, the player spawn is placed in the upper-left most section of the first segment placed, and the level ending orb is placed in the lower-right most section of the last segment placed.

\section{Experiments}

This section goes into detail regarding how levels were evolved and evaluated based on desirable properties. Both the OneGAN and MultiGAN methods use NSGA-II~\cite{deb:tec02} to evolve suitable levels. The two fitness objectives are solution path length and connectivity, both determined by A* search. Parameters for evolution are also explained in this section. Code for the experiments is available as part of the MM-NEAT software framework\footnote{\url{https://github.com/schrum2/MM-NEAT}}.

\subsection{NSGA-II}
NSGA-II~\cite{deb:tec02} is a Pareto-based multi-objective evolutionary optimization algorithm. NSGA-II uses the concept of Pareto Dominance to sort populations into Pareto layers based on their objective scores. If an individual is at least as good as another in all objectives, and strictly better in at least one objective, then that individual \emph{dominates} the other. The most dominant Pareto layer contains individuals which are not dominated by any other individual in the population, and is known as the nondominated Pareto front. 

NSGA-II uses elitist selection favoring individuals in the Pareto front
over others. The second tier of individuals consists of those that would be nondominated if the Pareto front were absent, and further layers can be defined by peeling off higher layers in this fashion. When ties need to be broken among individuals in the same layer, individuals that are maximally spread out to lower density regions of the current layer are preferred.

Pareto-based algorithms like NSGA-II are useful when objectives can be in conflict with each other, thus causing trade-offs. However, even when objectives correlate fairly well with each other, as in this paper, NSGA-II is useful in that it can provide a fitness signal in regions of the search space when one objective may be flat, without the need to define any kind of weighting between objectives. 

\subsection{Fitness Functions}

Levels are evolved with a combination of connectivity score and solution path length: the connectivity score provides a smooth gradient to improvement, even in regions of the search space filled with unbeatable levels whose solution path length is undefined.

Solution path length is determined by A* search 
on a simplified model of the game to allow for quick execution. If A* could not beat the level, then it was assigned a score of $-1$, and levels that were beatable were assigned a score equal to the length of the A* path. 


When multiple levels are unbeatable, connectivity  provides a way of differentiating them. Connectivity can also differentiate two levels with the same solution length. Connectivity score is formally defined as the proportion of traversable tiles (e.g.\ air, ladders) in the level that are reachable.
Higher connectivity implies more places to explore, thus containing less wasted/unused space.


The simplified A* model only allows the avatar to move discretely through the space one tile position at a time, and has a simplified jumping model that simulates the playable game. The model recognizes that Mega Man will die from contact with spikes or due to falling off the edge of the screen, but ignores the existence of enemies in the level. Though the model is not perfect, it is sufficient, and faster to calculate than an actual simulation of the full dynamics of Mega Man. To verify that all evolved champion levels were beatable, they were uploaded to the Mega Man Maker servers, which do not allow levels to be uploaded unless a human player successfully beats them first.

\subsection{Experimental Parameters}

Levels were evolved using both the OneGAN and MultiGAN approach using NSGA-II with a population size of $\mu = \lambda = 100$ for 300 generations. Preliminary experiments indicated no significant improvements after 300 generations. The evolution experiment was repeated with each pre-trained generator 30 times. 

Evolved levels consisted of 10 segments each, unless generation terminated early (see Section~\ref{sec:genome}). Since the GAN latent input size was 5, and each segment used 4 auxiliary variables for determining relative placement, the total length of each real-valued genome was $(5+4)\times 10 = 90$ variables in the range $[-1,1]$. Each generation offspring were created using a 50\% chance of single-point crossover. Whether crossover occurred or not, every offspring was mutated: each real-valued number in the vector
had an independent 30\% chance of polynomial mutation \cite{deb1:cs95:polynomial}.

\section{Results}

Levels produced by OneGAN and MultiGAN are compared quantitatively in terms of their solution path lengths and novelty, and qualitatively in terms of the final levels produced by each run. 

\subsection{Quantitative Analysis}
\begin{figure}[t]
\centering
\includegraphics[width=1.0\columnwidth]{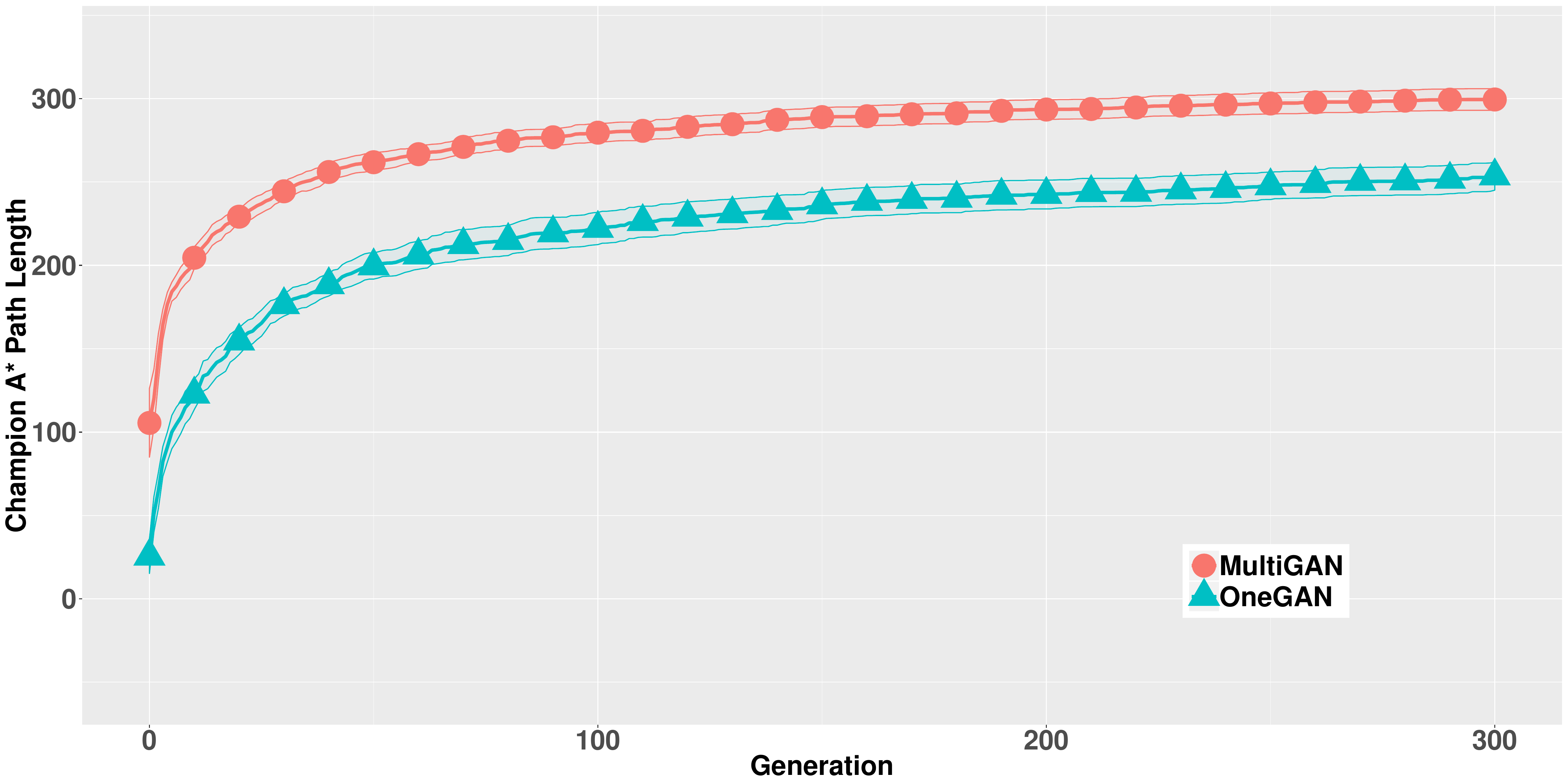}
\caption{Average Champion A* Path Length Across Evolution. \normalfont Plot of averages across 30 runs of OneGAN and MultiGAN of champion A* path lengths by generation. MultiGAN path lengths are approximately 50 steps longer than OneGAN paths throughout evolution. For context, recall that segments are $14 \times 16$ tiles.}
\label{fig:averageScores}
\end{figure}

The A* path lengths are significantly longer ($p < 0.05$) with MultiGAN than OneGAN (Fig.~\ref{fig:averageScores}). The separation between methods is established early in evolution and maintained until the end. The difference in averages is approximately 50 tile steps throughout all of evolution. Because segments are $14 \times 16$ tiles, a difference of 50 means that OneGAN champions sometimes skip one or more segments, though it is also possible for paths to be lengthened with additional twists and turns inside individual segments.

Levels produced by the two approaches were also compared in terms of a novelty metric. Novelty can be defined in terms of an individual segment, a whole level, or an arbitrary collection of segments. Segment distance $d(x,y)$ between two segments $x$ and $y$ is defined as the number of positions in which their respective tile types differ, normalized to $[0,1]$. Then, the novelty $N$ of a segment $x$ is defined as its average distance from all segments in some reference collection $S$:
\begin{equation}
    N(x,S) = \frac{\sum_{y \in S} d(x,y)}{|S|}
\end{equation}
The novelty of a level $M$ is the average novelty of all segments it contains, where each segment's novelty is calculated with respect to the set of other segments in the level (excluding itself). The Level Novelty $LN$ is:
\begin{equation}
    LN(M) = \frac{\sum_{x \in M} N(x, M - \{x\})}{|M|}
\end{equation}


Within VGLC levels, segments are fairly uniform and less novel. Both GAN approaches produce levels where there is greater segment variety in each level compared to VGLC (Table~\ref{tab:avgnovelty}). This suggests that there is more variety in GAN-generated levels than in the original game, but it may also suggest less consistent style. 

Table~\ref{tab:setnovelty} indicates that there are not that many repeated segments within the individual data sets for a given type of level. MultiGAN repeats segments more than OneGAN (fewer unique segments), likely because of the lack of training samples in corner segment GANs. 
In fact, Table~\ref{tab:cornernovelty} analyzes the distinct segments and novelty scores of each corner GAN from MultiGAN, showing that these corner GANs produced less novel results. However, despite some repetition in the corners, the overall novelty scores of OneGAN and MultiGAN are slightly more than VGLC. 



\begin{table}[t!]
\caption{\label{tab:avgnovelty}Average Level Novelty By Type}
{\small For each type of level, the $LN$ of each level is calculated. The
average $LN$ score across $N$ levels of the given type are presented below.}

\centering
\begin{tabular}{|c|c|c|c|c|}
\hline
Type & $N$ & Average $\pm$ StDev & Min & Max \\ 
\hline
VGLC& 10 & 0.34$\pm$0.09 & 0.11 & 0.45\\ 
\hline 
OneGAN & 30 & 0.43$\pm$0.03 & 0.36 & 0.48\\
\hline
MultiGAN & 30 & 0.41$\pm$0.02 & 0.34 & 0.46 \\
\hline

\end{tabular}
\end{table}

\begin{table}[t!]
\caption{\label{tab:setnovelty} Distinct Segments By Type}
{\small For each level type, the total number of segments,
number and percentage of unique segments within the collection (removing duplicates to get a set),
average segment novelty with respect to the collection, and average segment novelty with respect to the set (without duplicates) are shown.
MultiGAN has the lowest percentage of
distinct segments, but is between VGLC and OneGAN in terms of novelty, whether sets or complete collections are used, though the comparative novelty scores are all close.}


\centering
\resizebox{\columnwidth}{!}{
\begin{tabular}{|c|c|c|c|c|}
\hline
 & Segments & Distinct       & Average          & Average\\ 
 &          & Segments       & Novelty All      & Novelty Set\\
\hline 
VGLC & 178  & 159 (89.3\%) & 0.4390 & 0.4349  \\
\hline
OneGAN & 300 & 287 (95.7\%) & 0.4709  & 0.4637  \\
\hline
MultiGAN & 300 & 254 (84.7\%) & 0.4542 & 0.4483\\
\hline

\end{tabular}
}
\end{table}





\begin{table}[t!]
\caption{\label{tab:cornernovelty} Distinct Corner Segments in MultiGAN}
{\small Shows number of segments in MultiGAN levels generated by each corner GAN, number that were distinct, average novelty of collections from each GAN, and average novelty across the distinct segments. Corner segments from the same GAN often differ by only a few tiles, which is why novelty scores are low despite the sometimes high percentage of distinct segments.}
\centering
\resizebox{\columnwidth}{!}{
\begin{tabular}{|c|c|c|c|c|}
\hline
 & Segments & Distinct & Average & Average\\ 
 &          & Segments       & Novelty All      & Novelty Set\\
\hline 
Lower Left & 42 & 28 (66.7\%) & 0.2453 & 0.2689 \\
\hline
Lower Right & 25 & 20 (80.0\%) & 0.3537 & 0.3448 \\
\hline
Upper Right & 40 & 29 (72.5\%) & 0.2635 & 0.2770 \\
\hline
Upper Left & 25 & 22 (88.0\%) & 0.2875 & 0.2857 \\
\hline
\end{tabular}
}
\end{table}




\setlength\tabcolsep{1.0pt}
\begin{table*}[ph!]
\caption{\label{tab:LEVELS} Example Generated Levels.}
{\small Each is discussed in detail in Section~\ref{sec:levelapp}. Names appear under each level.}

\centering
\begin{tabular}{p{0.35\textwidth}p{0.4\textwidth}}
 \multirow{6}{0.3\textwidth}[4.5cm]{
 \includegraphics[width=0.3\textwidth]{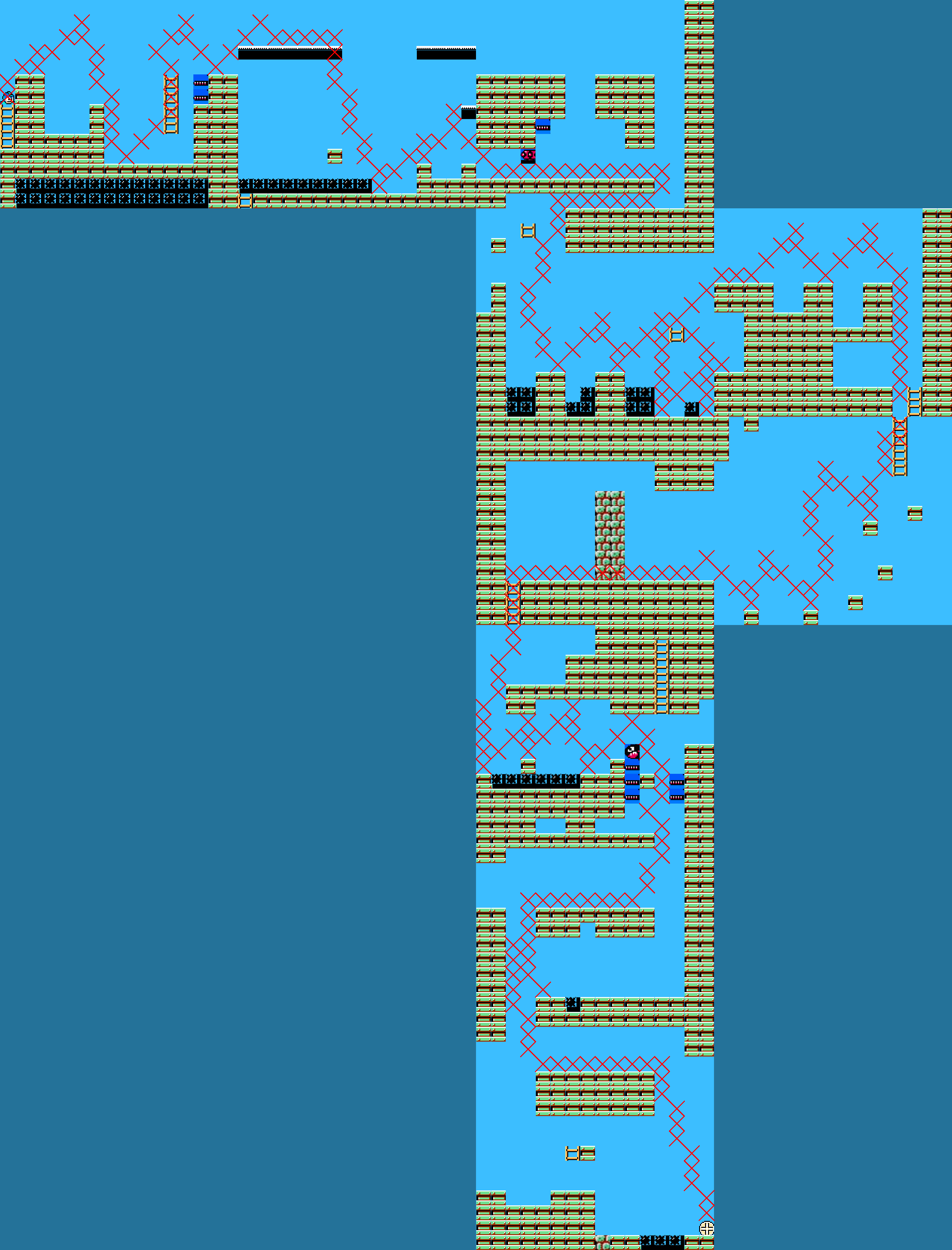} 
 MultiGAN1 
 \includegraphics[width=0.25\textwidth]{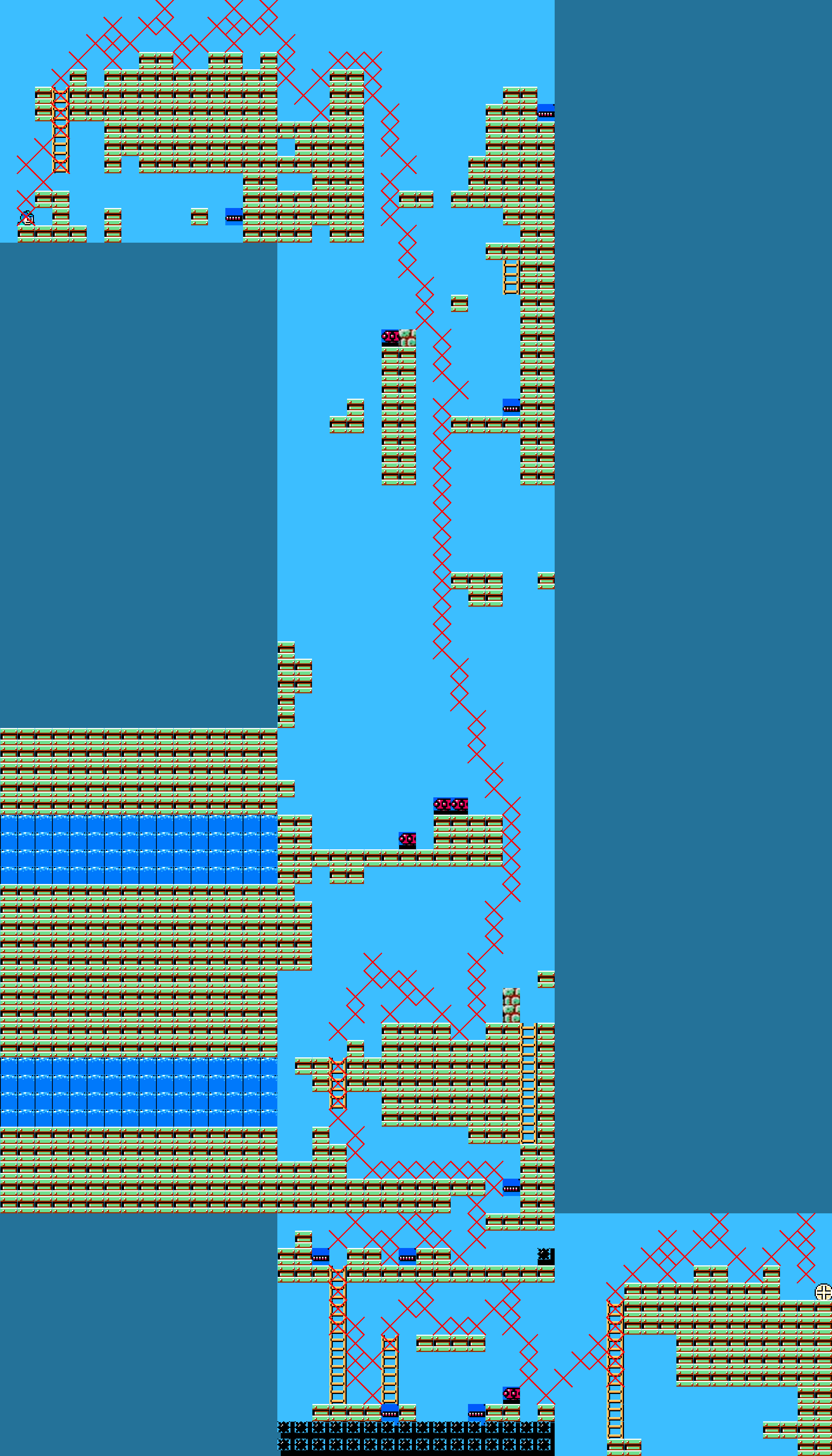} 
OneGAN15} & 
\includegraphics[width=0.35\textwidth]{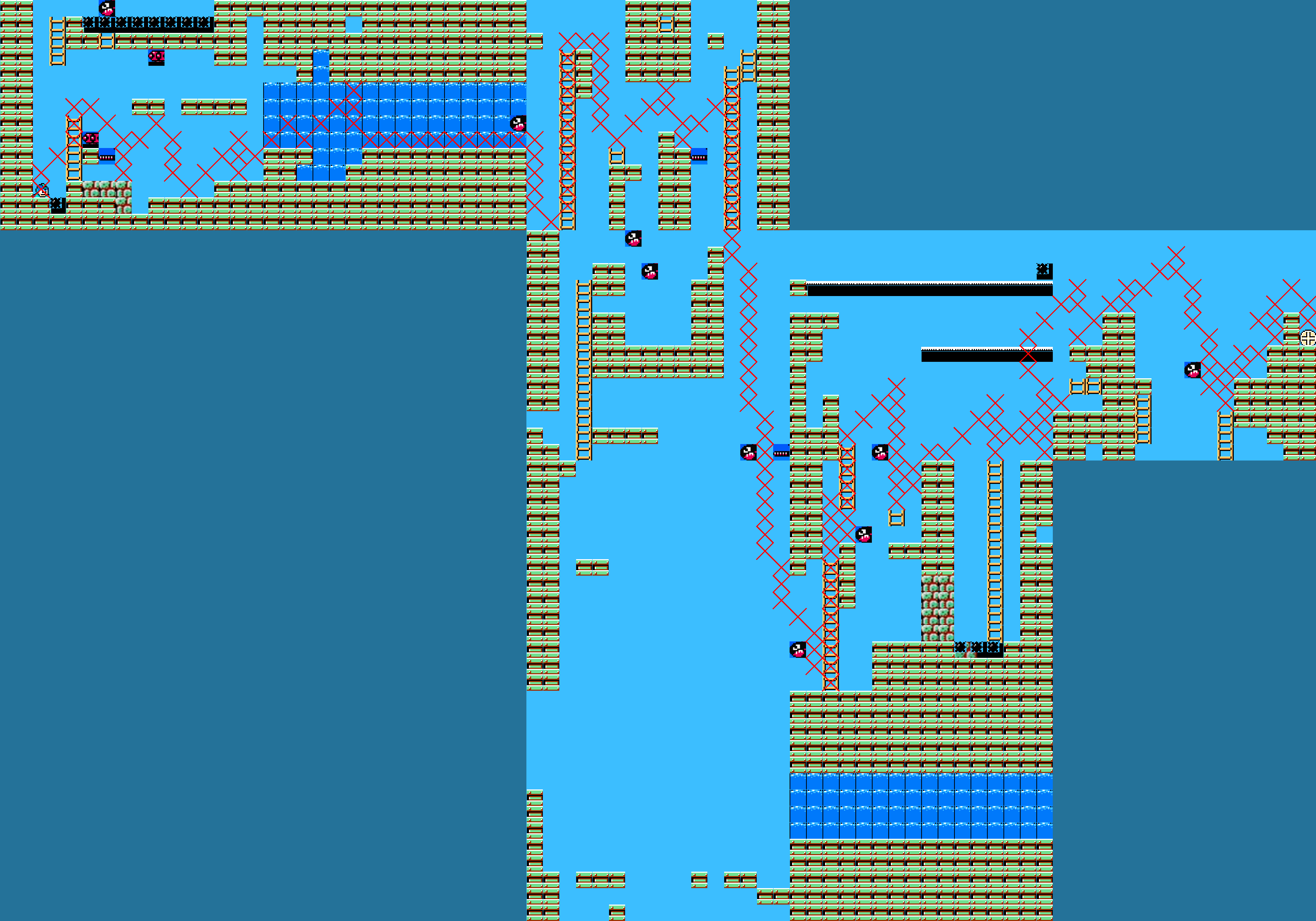} \\
& OneGAN0\\ 
& \includegraphics[width=0.35\textwidth]{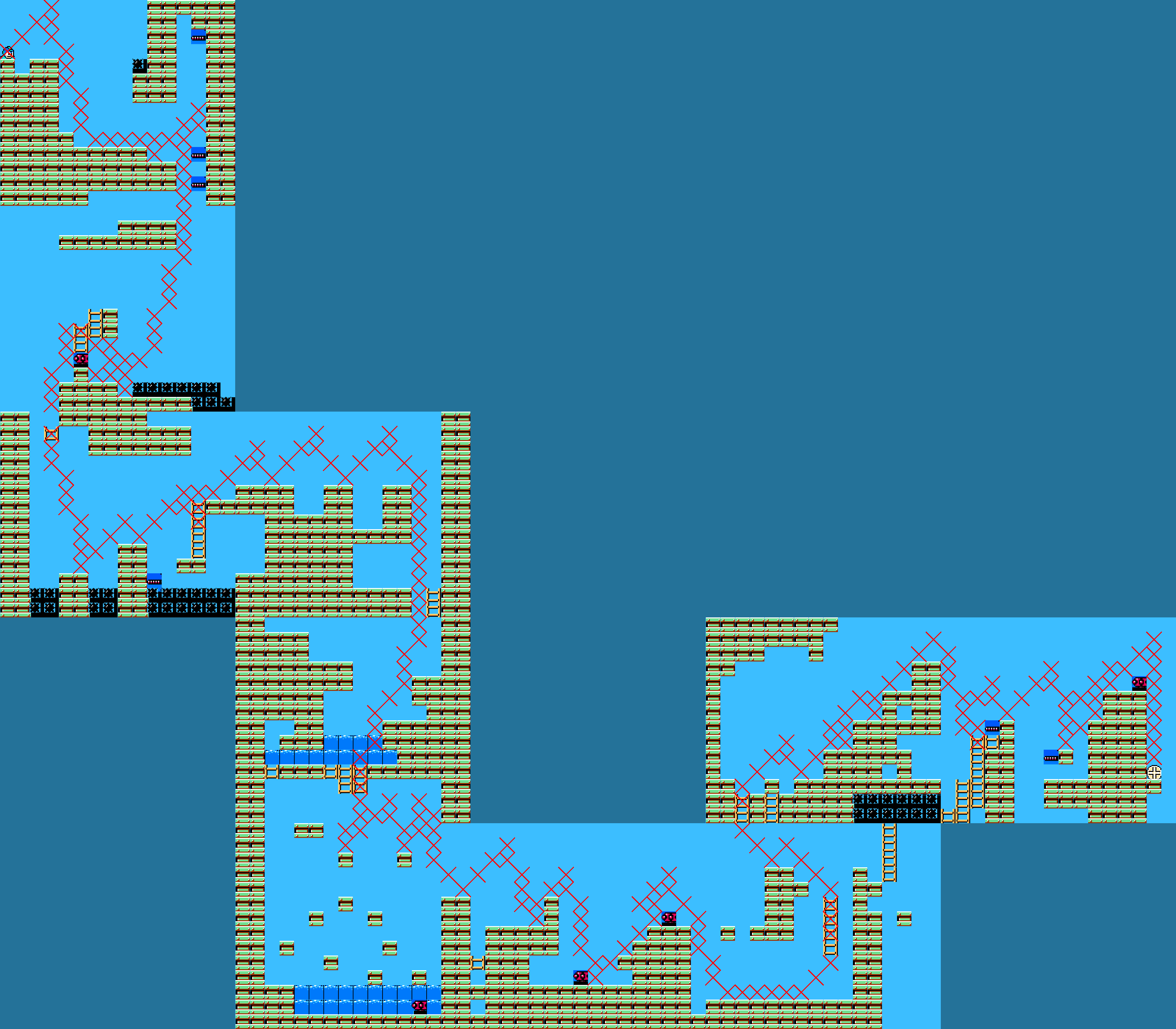}\\
& MultiGAN25\\ 
& \includegraphics[width=0.3\textwidth]{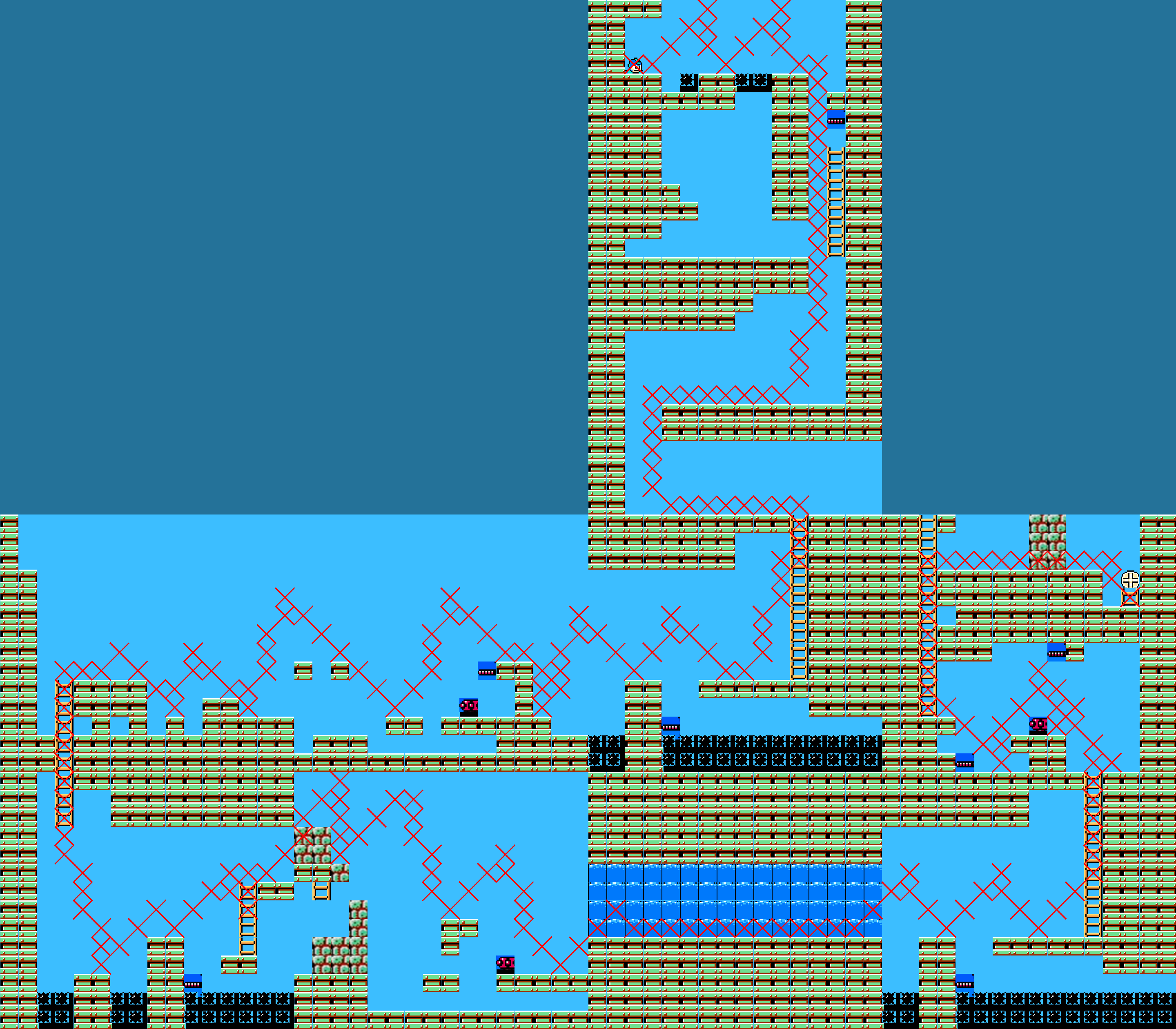}\\
& MultiGAN27\\

\includegraphics[width=0.35\textwidth]{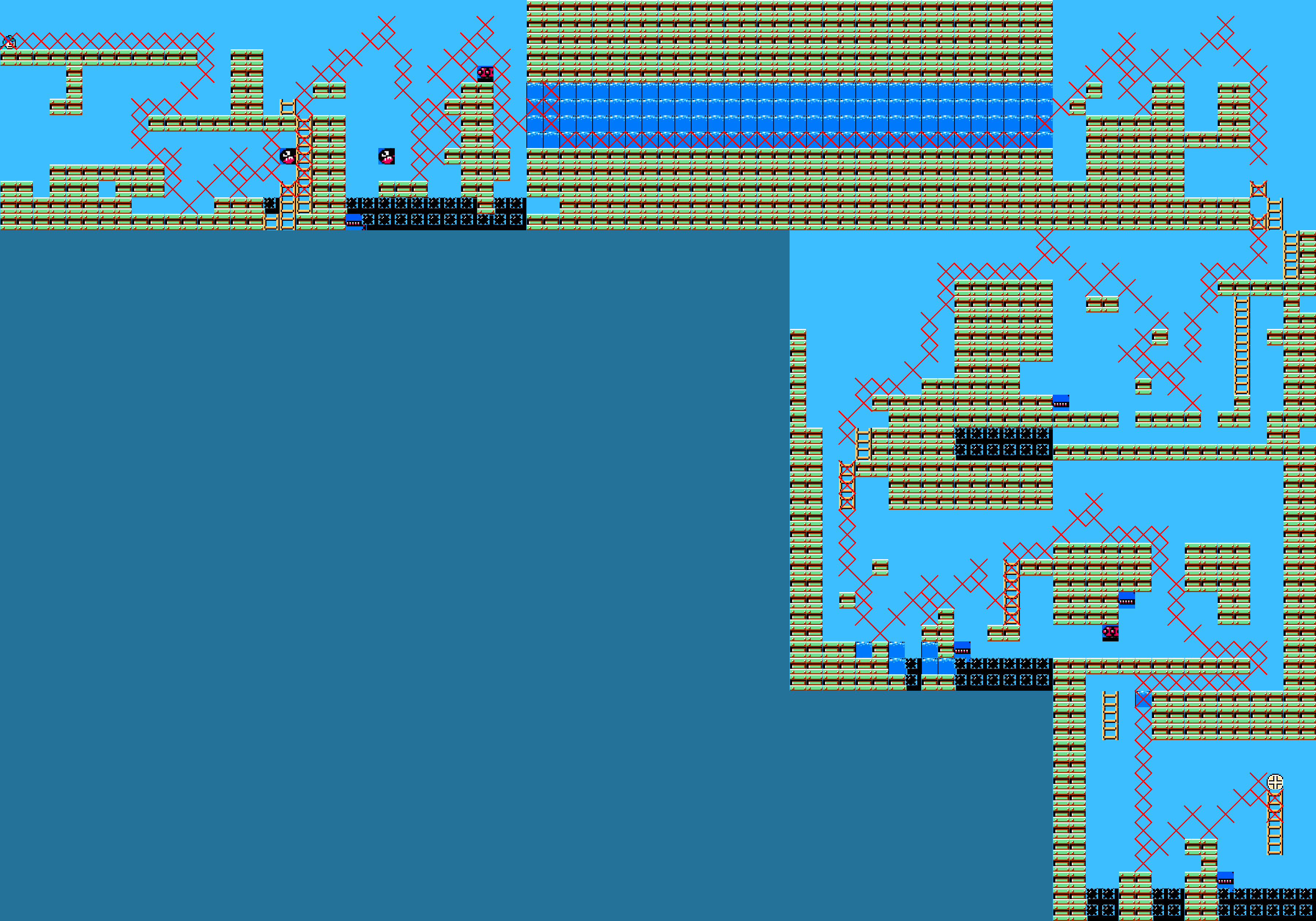}
MultiGAN7
& \includegraphics[width=0.4\textwidth]{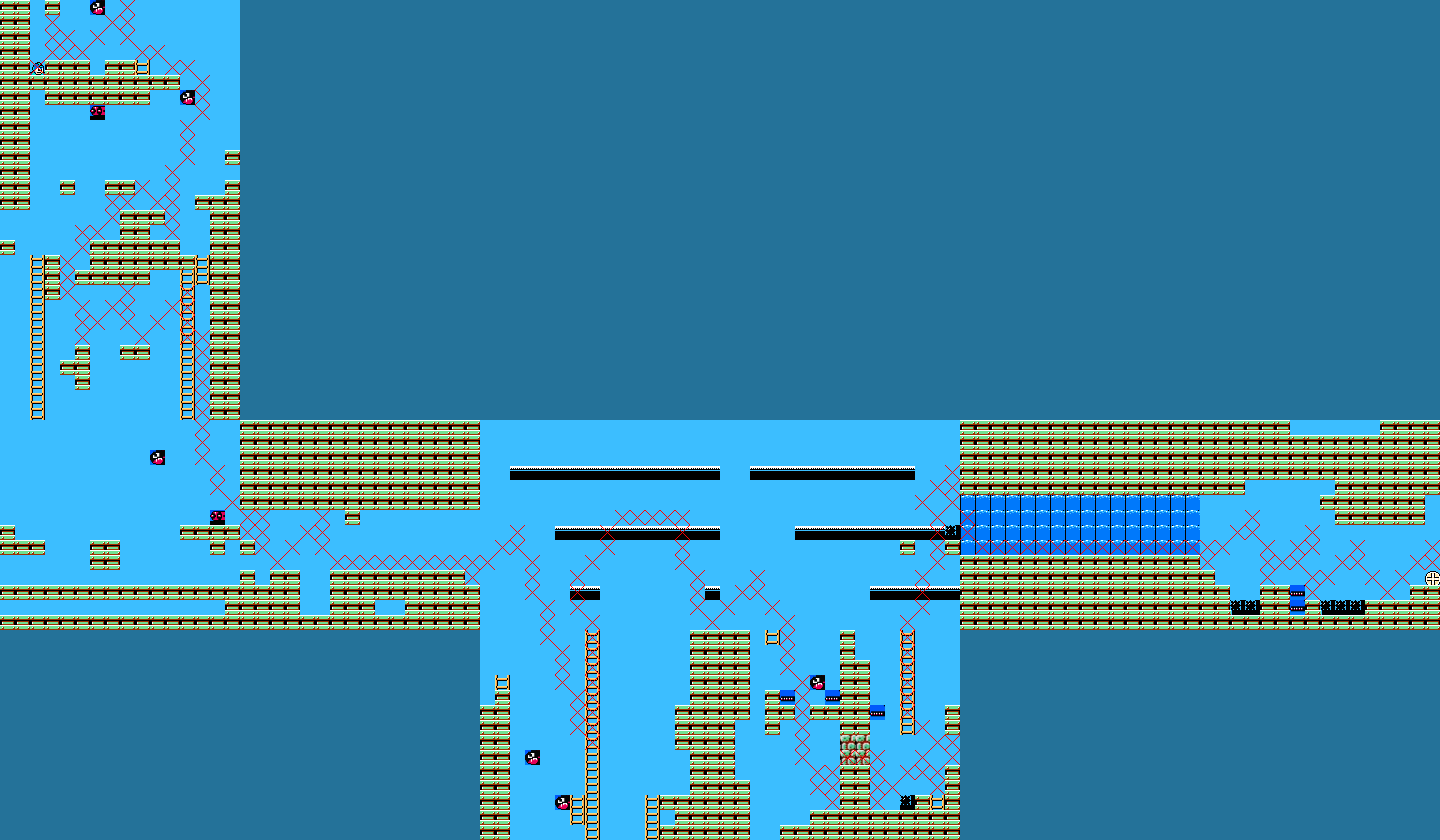} 
OneGAN8\\

\end{tabular}
\end{table*}

\subsection{Qualitative Analysis}
\label{sec:levelapp}

The specific levels talked about in this section can be found in Table~\ref{tab:LEVELS}. Evolved champions from all 30 runs with OneGAN and MultiGAN can be viewed online\footnote{\url{https://southwestern.edu/~schrum2/SCOPE/megaman.php}} and played with Mega Man Maker.

OneGAN levels were more confusing due to the lack of flow. There are many random pits that Mega Man cannot traverse without the platform gun (see Section 3). MultiGAN levels generally look more natural, meaning adjacent segments connect and flow better. 


Different MultiGAN levels tend to have the same or nearly identical corner segments because of
issues with novelty pointed out in Table \ref{tab:cornernovelty}. However, a typical 10-segment level often only has one or two representatives of each corner type, so repetition of corner segments within the same level is very rare.





As seen in \textbf{OneGAN15}, OneGAN levels tend to have entire segments that are unreachable or unnecessary. One water segment in the figure 
leads to a dead end, and the other is blocked off by the other segments. Note that despite being specifically evolved to maximize
solution length and connectivity, OneGAN struggled to connect segments. Similarly, in \textbf{OneGAN0} there are two unused segments at the bottom.
These segments are reachable, but not needed. Examples like these explain why OneGAN's solution path lengths are on average 50 tiles shorter than those of MultiGAN.


However, OneGAN solutions do sometimes successfully traverse all segments, as seen in \textbf{OneGAN8}. However, even this level has large sections in the lower left and upper mid section that are not traversed, which further explains the shorter solution paths compared to MultiGAN.




In contrast, MultiGAN levels 
made better use of their allotment of 10 segments.
In \textbf{MultiGAN1}, the average portion of each segment occupied by the solution path is roughly $12\%$, whereas in \textbf{OneGAN15} the portion is only $8\%$. Every ladder in the level is part of the solution path, suggesting a more efficient use of space and intelligent placement. In fact, the left-most side of \textbf{MultiGAN27} even presents an example of ladders between distinct segments perfectly aligning.
\textbf{MultiGAN1} also effectively paired moving platforms with hazard spikes, which is an interesting challenge. MultiGAN levels were good at placing blocks in such a way that a simple miscalculated jump could result in death, as in the original Mega Man, which is known for frustrating platforming challenges. 

MultiGAN levels generally follow a snaking pattern not only in segment placement, but also in the solution path, whereas solution paths for OneGAN levels were more direct. Though MultiGAN does occasionally struggle with unnecessary block placement, such as in \textbf{MultiGAN25} with the lower left corner, the solution paths are both longer and more challenging to traverse than in OneGAN.

Human subjects confirmed some of these observations.

\section{Human Subject Study}

This section briefly describes a human subject study comparing OneGAN and MultiGAN levels, and the results of the study.

\subsection{Human Subject Study Procedure}

The study was advertised globally on social media and forums, but most of the 20 participants were undergraduate students from Southwestern University.
Response was limited because participants seemed unwilling or unable to install and create an account for Mega Man Maker, which was required in order to participate.


Participants played a random evolved champion level of each type. Subjects were not informed how levels were generated, and were led to believe that some levels \emph{might} be made by humans. Participants compared levels based on how difficult, fun, human-like, and interesting the designs were.





Mega Man Maker allows for a rich diversity of tiles that vary mainly in their appearance, but the tile types produced by the GAN (Table \ref{tab:tiles}) are relatively plain. As a result, levels look simpler than typical human-designed levels.
One study participant who had experience with Mega Man Maker pointed out that our levels would be impossible to make using the level editor because the editor blends different tiles from a given tile set to distinguish surfaces, corners, etc. As a result, this user could tell that our levels were made through direct manipulation of level text files.

Players had access to the platform-gun, allowing them to more easily traverse the more difficult, or otherwise impossible, jumping puzzles. Participants were asked if they thought the platform-gun was required to beat the level. Even with the platform-gun, some players did not beat both levels. They were encouraged, though not required, to make several attempts at each.



\subsection{Human Subject Study Results}


Quantitative results from the human subject study are in Table~\ref{tab:subjectStudyData}, including a statistical analysis.

Because it was difficult to find expert Mega Man players, several participants struggled to complete the levels.
Only 14 participants completed the OneGAN level, whereas 17 completed the MultiGAN level. Additionally, 11 thought the platform-gun was required to beat the OneGAN level, whereas only 8 believed this of the MultiGAN level. Needing the platform-gun could indicate bad design, as most actual levels can be traversed by normal jumping.



\begin{table}[t!]
\caption{\label{tab:subjectStudyData} Human Subject Study Results}
{\small Shows the data associated with the Human Subject Study that was conducted. Under Type, ``y/n'' means participants could answer \emph{yes} or \emph{no} for each individual level, and ``e/o'' means the participant had to pick \emph{either} one level \emph{or} the other. Responses to ``e/o'' questions were compared using one-sided binomial tests whose approximate $p$ values are shown in the last column, where \textbf{bold} values indicate a significant difference ($p < 0.05$)} \\
\centering
\begin{tabular}{|c|c|c|c|c|}
\hline
Question & Type & OneGAN & MultiGAN & $p$ \\ 
\hline 
\hline
Completed? & y/n & 14 & 17 & N/A \\
\hline
Platform-Gun Required? & y/n & 11 & 8 & N/A  \\
\hline
Created by a Human? & y/n & 4 & 13 & N/A \\
\hline
\hline
Harder? & e/o & 14 & 6 & 0.05766 \\
\hline
More Fun? & e/o & 2 & 18 & \textbf{0.0002012} \\
\hline
More Human-Like? & e/o & 5 & 15 & \textbf{0.02069}\\
\hline
More Interesting? & e/o & 7 & 13 & 0.1316\\
\hline


\end{tabular}
\end{table}


Most participants found their OneGAN level harder than their MultiGAN level. One respondent who said the OneGAN level was harder noted that the ladder placement ``made no sense/led to nowhere/death screen.'' Another respondent said the OneGAN level ``seemed hard in a random and not very well thought-out way.'' Yet another said they could not see where they were falling and that it was difficult to ``time [their] falls.'' A different participant said that OneGAN was harder because it ``seemed to force the player to make a leap of faith at the start'' which led to confusion.


A significant number of participants found the MultiGAN level more fun ($p < 0.05$). 
One participant said that the level ```knew' what [they were] going to do and could make things deliberately harder for [them] in an intelligent way,'' whereas the OneGAN level was ``hard in a seemingly unintentional way.'' Another participant said that they ``liked how much longer [MultiGAN] was. There were many places [they] could go.'' However, two participants said that neither were particularly enjoyable, and another participant said that the MultiGAN level ``felt more like [they] were supposed to lose, where [the OneGAN level]...wanted to be beaten.'' For that particular user, the OneGAN level had five segments where the player could fall down to quickly and easily traverse a large portion of the level, whereas the SevenGAN level had more enemies, harder jumping puzzles, and a longer solution path. Such comments also explain why a majority of users thought MultiGAN levels were more interesting than OneGAN levels.



A significant number of participants ($p < 0.05$) thought the MultiGAN level's design was more human-like than OneGAN's, and 13 thought the MultiGAN level was created by a human in comparison to OneGAN's 4, though some thought correctly that neither was designed by a human.
One participant said the OneGAN level was not created by a human because it spawns the player next to enemies, and it ``feels impossible to not take damage from them.'' Another participant said that the OneGAN level ``seemed like it was entirely random'' and that it made it ``frustrating.'' Yet another said that the MultiGAN level was made by a human because it ``had more repeated building elements'' and that the OneGAN level had no such pattern, and would sometimes have a ``solitary floating block.'' Finally, one participant said that the OneGAN level ``seemed more random'' than MultiGAN, which led to them thinking that it was ``made by a human trying to trick the player.''


\section{Discussion and Future Work}


With OneGAN, the barriers between segments are unpredictable. The majority of the training data consists of horizontal segments, making such segments more likely. However, if Mega Man needs to move up into a new segment, he could get stuck banging into the floor of the segment above him. Often, when Mega Man needs to move down to a new segment, the only way to do so is by falling into what looks like a pit trap, because the next segment was generated beneath a segment modelled on horizontal segments. Note that OneGAN does not appear to be suffering from mode collapse~\cite{thanhtung2020catastrophic}; but it does struggle to pick the right type of segment for a given situation. Because there are higher odds of the next segment being unreachable, the only way to traverse some OneGAN levels is by side routes caused when the sequence of segments snakes back into being adjacent to an earlier segment. However, these side routes result in certain segments being skipped in the solution path.

MultiGAN does not have this problem. When an upward segment is needed, it can be generated reliably. Everything generated by the Up GAN will provide ladders and/or platforms that make such movement possible. Similarly, when movement transitions from horizontal to vertical, the MultiGAN will use an appropriate corner GAN.
However, corner GANs are trained on significantly fewer segments than their horizontal and vertical counterparts, causing a lack of diversity in generated corner segments (Table \ref{tab:cornernovelty}). This lack of diversity will generally not be noticed in any individual MultiGAN level, but repetition of specific corner segments can be seen across levels. 
Also, those familiar with the original game may notice the duplication of certain corner segments from Mega Man.


Both OneGAN and MultiGAN sometimes produce plain hallways filled with water.
This segment is a reproduction of a segment that occurs repeatedly in Level 9, which consists of a long boring hallway filled with water. In the context of the original game, this scenario is an interesting departure from the norm, but the appearance of this segment in levels produced by GANs is usually out of place. In fact, the handling of water by both approaches can lead to unusual segments on occasion, indicating that some special handling of segments containing water may be necessary. However, despite the lack of continuity when handling water segments, LVE resulted in other types of segments derived from the diverse training set fitting well together.


Conditional Generative Adversarial Networks (CGANs \cite{mirza2014conditional}) could serve as an alternative to using multiple GANs. A CGAN is conditioned on some additional input, allowing it to produce output of a desired type on demand, thus eliminating the need for multiple GANs. 
However, the imbalance of training data would be a more serious issue for CGANs than it is for MultiGAN. The data disparity between horizontal/corner segments is nearly 150 to 1.



Larger training sets could solve these problems.
There are many possible sources for Mega Man levels, such as the many other games in the Mega Man franchise. 
Data for games beyond Mega Man 1 are not in VGLC, but such a data set could be constructed. A more readily available source of levels is on the Mega Man Maker servers. These can be freely downloaded, and if properly simplified, could be converted into a format usable for training.
With a large enough training set, MultiGAN should be able to produce greater diversity in its corner segments. However, more data in general means there will still be an imbalance with respect to corner segments, so simply using more data still might not make CGANs easier to apply.

In lieu of more training data, better corner diversity could be achieved by being more permissive about what counts as a corner segment. Currently, a sequence of horizontal and vertical window slices lead into and out of every occurrence of a corner segment. However, the three or four window slices surrounding each corner segment could potentially serve as viable corner segments as well. For example, if a corner segment has a floor section that is four tiles deep, then sliding the window up three tiles still leaves a floor at the bottom of the screen. Ladders and platforms for vertical movement would remain. Including such segments in the training sets for corners would increase their sizes without resulting in corner segments that could not be traversed. These new members of the corner sets could also be removed from horizontal and vertical data sets, slightly improving the balance of data. 



A larger problem affecting OneGAN and, to a lesser extent, MultiGAN, is the issue of continuity between adjacent segments. The levels in the training set are all stylistically different,
so if one simply stitched together a series of horizontal segments from different levels, the result would likely not be cohesive. 
To some extent, use of proper fitness functions during evolution creates some cohesion, but it would be easier if evolution did not need to make up for shortcomings in the GAN and genotype encoding.

A recently developed approach that could address this issue is CPPN2GAN \cite{schrum:gecco2020cppn2gan}, which uses Compositional Pattern Producing Networks (CPPNs~\cite{stanley:gpem2007}) to generate GAN latent inputs. CPPN outputs vary gradually as a function of segment location within the level, and since similar latent vectors result in similar GAN outputs, adjacent segments should be linked in a more cohesive way. This approach has the added benefit of scale, because levels of arbitrary size can be generated by a compact CPPN.


Because MultiGAN produced less variety in its corner segments, its levels were slightly less diverse than OneGAN's. Having better training data could fix this, but another way to increase diversity is to explicitly favor it during evolution. Quality diversity algorithms like MAP-Elites~\cite{mouret:arxiv15} could help in discovering such diverse levels, as has already been done in other GAN-based approaches \cite{schrum:gecco2020cppn2gan,fontaine2020illuminating,steckel2021illuminating}.



\section{Conclusion}

When using GANs to generate levels with various segment types, it helps to use multiple GANs. Doing so preserves the structure of each segment type. In Mega Man, each segment type affects how adjacent segments connect.
If poorly connected segments are generated, as with OneGAN, then Mega Man cannot properly traverse the entire level.
MultiGAN is proposed to allow generation of the right type of segment whenever needed. 
MultiGAN was effective in producing levels with longer solution paths going through all available segments, in a way reminiscent of human-designed levels from the original game. 
In fact, a significant number of human subjects confirmed that MultiGAN levels had more human-like design and were more fun, in contrast to OneGAN levels which often had unusual barriers, unreachable segments, and overall stranger level design. 
MultiGAN shows promise for the generation of more complex, challenging, and cohesive levels, and future extensions to the approach should result in more diverse level designs as well.


\begin{acks}

This research was made possible by the donation-funded Summer Collaborative Opportunities and Experiences (SCOPE) program for undergraduate research at Southwestern University

\end{acks}

\bibliographystyle{ACM-Reference-Format}
\bibliography{MegaMan} 


\end{document}